\documentclass{article} 
\usepackage{iclr2024_conference, times}
\usepackage{multirow}
\usepackage{tabularx}
\usepackage{xcolor} 
\usepackage{graphicx} 
\usepackage{tcolorbox}
\usepackage{listings}
\usepackage{amsmath,amsfonts,bm}

\def\eqref#1{equation~\ref{#1}}

\def\1{\bm{1}}

\DeclareMathAlphabet{\mathsfit}{\encodingdefault}{\sfdefault}{m}{sl}
\SetMathAlphabet{\mathsfit}{bold}{\encodingdefault}{\sfdefault}{bx}{n}

\usepackage{hyperref}
\usepackage{url}

\title{Safurai 001: New Qualitative Approach for Code LLM Evaluation}

\author{Cifarelli D., Boiardi L. \& Puppo A. \thanks{https://www.safurai.com/team} \\
Safurai Team\\
Genoa\\
Corso Firenze 39/3, 16136, Italy \\
\texttt{info@safurai.com} \\
}

\iclrfinalcopy 
\begin{document}

\maketitle

\begin{abstract}
This paper presents Safurai-001, a new Large Language Model (LLM) with significant potential in the domain of coding assistance. Driven by recent advancements in coding LLMs, Safurai-001 competes in performance with the latest models like WizardCoder [\cite{wizardcoder}], PanguCoder [\cite{pangucoder}] and Phi-1 [\cite{phi-1}] but aims to deliver a more "conversational" interaction. By capitalizing on the progress in data engineering (latest techniques of data transformation and prompt engineering) and instruction tuning, this new model promises to stand toe-to-toe with recent closed and open source developments. Recognizing the need for an efficacious evaluation metric for coding LLMs, this paper also introduces GPT4-based MultiParameters:  an evaluation benchmark that harnesses varied parameters to present a comprehensive insight into the model's functioning and performance. Our assessment shows that Safurai-001 can outperform GPT-3.5\footnote{https://openai.com/blog/introducing-chatgpt-and-whisper-apis} by 1.58\% and WizardCoder by 18.78\% in Code Readability parameter and more. 
\end{abstract}

\section{Introduction}
Code large language models are one of the most promising applications of LLMs and they have drawn a lot of interest from both academia and industry because of their extraordinary aptitude for tasks involving codes.

The closed-source models landscape is dominated by OpenAI models:  GPT-3.5 and GPT-4 [\cite{gpt4}] (actually, the best ranked model in HumanEval pass@1 chart). Before the release of Starcoder [\cite{starcoder}], open-source world fall far behind commercial models in terms of model size, capability, and performance.

However, this paradigm started changing with the advent of Starcoder. It have been frequently employed as a foundational model in the development of other models with great results like WizardCoder [\cite{wizardcoder}] and PanguCoder [\cite{pangucoder}], diminishing significantly the performance gap between open and closed-source coding LLMs.
Lately, also Meta introduced a new set of 12 LLMs available for commercial use: LLAMA2 release [\cite{llama2}]. 
Teams from all over in the world could use LLAMA2 as new foundation model for Coding LLMs, in competition with StarCoder.

In the latest publications in Coding LLMs field, many efforts have been made regarding for data engineering (Phi-1) and instruction tuning (WizardCoder). 

We have tried to capitalize on all the latest innovations in the field of Coding LLMs to develop a high-performance model that is in line with the latest open-source releases.

In a nutshell, we make the following contribution:
\begin{itemize}
    \item We present Safurai-001, a model that competes with WizardCoder for performances and tries to have a more "conversational" approach.
    \item We introduce a new evaluation benchmark for coding LLMs, GPT4-based MultiParameters Evaluation Benchmark. This benchmark embraces multiple crucial parameters to offer deeper insights into the model's performance.
\end{itemize}

\section{Related Work}
\subsection{Coding Large Language Models}
The impressive Codex model, with its 12 billion parameters, illustrates a remarkable capacity to solve approximately 72\% of Python programming challenges. This achievement has paved the way for the development of other advanced code generation models, including AlphaCode [\cite{alphacode}], PaLM-Coder [\cite{palm}], and PanGu-Coder [\cite{pangucoder}]. However, one notable drawback is the lack of open-source availability of these state-of-the-art models, a void that has subsequently been filled by the release of several open-source variants such as CodeParrot\footnote{https://huggingface.co/codeparrot/codeparrot}, PolyCoder\footnote{https://huggingface.co/NinedayWang/PolyCoder-2.7B}, PyCodeGPT\footnote{https://github.com/microsoft/PyCodeGPT}, SantaCoder [\cite{santacoder}], and StarCoder [\cite{starcoder}]. This new wave of open-source models have reinvigorated the code generation field.

Furthermore, the sequential expansions of code generation application scopes are reflective of the field's ever-growing practicality. For instance, CodeGeeX [\cite{codegeex}], BLOOM [\cite{bloom}] and ERNIE-Code [\cite{erniecode}] have been developed to enable multilingual modeling. JuPyT5 [\cite{jupyt5}] was trained using an extensive corpus of Jupyter notebooks, its primary objective being to enhance the process of interactive programming. Models like DocCoder and APICoder [\cite{apicoder}] have also been constructed to equip language models with the functionality to call APIs. Moreover, a number of models, including InCoder [\cite{incoder}], SantaCoder, and StarCoder, support code generation at arbitrary locations.

Recently, some groups have been utilizing instructional tuning techniques to tap into the vast potential knowledge contained within extensive language models. This process involves carefully refining these models with high-quality datasets. In terms of code generation, WizardCoder (15B), PanguCoder and phi-1 (1.3B) models stand out for their exemplary performance. This was achieved through careful fine-tuning with data generated by OpenAI's GPT-3.5 and GPT-4.

\subsection{Code, Algebra and Logic Dataset Landscape}

The landscape of code, logic, and algebra datasets is teeming with new possible resources that can be used for finetuning Coding LLMs (the majority of them are open source). 

The most important coding dataset in this field is CodeAlpaca-20k\footnote{https://huggingface.co/datasets/HuggingFaceH4/CodeAlpaca\_20K}. Many models, like PanGu-Coder or WizardCoder, have structured their dataset also through the manipulation of Code Alpaca with data augmentation techniques.
Also Phi-1 [\cite{phi-1}] coding model has been trained with filtered code-language dataset, which is a subset of The Stack\footnote{https://huggingface.co/datasets/bigcode/the-stack} (it contains over 6TB of permissively-licensed source code files covering 358 programming languages). 

The open source community offers a variety of resources in Q\&A format that are helpful for fine-tuning LLMs in terms of datasets for mathematics and logic. The majority of these datasets were produced by T5\footnote{https://huggingface.co/docs/transformers/model\_doc/t5}, GPT-3.5, GPT-4, or a combination of these models (although OpenAI policies can still be interpreted in this context).

\subsection{Latest Techniques for Prompt Engineering}
In this section, we outline the primary prompt engineering methods combined with prompt engineering, applied to the coding LLMs field:
\begin{itemize}
    \item Chain of Thoughts (CoT): Wei et al.[2023] report that large language models can enable the emergence of reasoning abilities when prompted in this way. A chain of thought is a series of intermediate natural language reasoning steps that lead to the final output.
    \item CoT and Self-Consistency: this is the natural evolution of CoT technique. It first samples a diverse set of reasoning paths instead of only taking the greedy one, and then selects the most consistent answer by marginalizing out the sampled reasoning paths. Self-consistency leverages the intuition that a complex reasoning problem typically admits multiple different ways of thinking leading to its unique correct answer (Wang et al.[2022]).
    \item Tree of Thoughts (ToT): Yao et al.[2023] report that ToT allows LMs to perform deliberate decision making by considering multiple different reasoning paths and self-evaluating choices to decide the next course of action, as well as looking ahead or backtracking when necessary to make global choices.
    \item Teacher CoT: Ho et al.[2023] demonstrated that through the augmentation of the prompt with an "educational" explanation generated by a larger model, excellent results are obtained in the finetuning of smaller models. Also Mukherjee et al.[2023] used this "teaching" approach to develop Orca model. 
    \item EvolInstruct: Luo et al.[2023] proposed a new approach for data augmentation that achieved important results. They found that LLMs can make given instructions more complex and difficult using specific prompts. Additionally, models can generate entirely new instructions that are equally complex but completely different. Using this discovery, the WizardCoder creators can iteratively evolve an initial instruction dataset, improving difficulty level and expanding its richness and diversity
\end{itemize}

\subsection{Latest Evaluation Techniques for Coding LLMs (HumanEval, MBPP, MultiPL-E, HumanEval Pack)}

This subchapter provides an overview of the benchmarks currently being used to evaluate Coding LLMs. 
\begin{enumerate}
    \item HumanEval\footnote{https://huggingface.co/datasets/openai\_humaneval}: This general standard benchmark holds a set of 163 problems constrained to Python language. It assesses whether the model’s code successfully passes all the tests and provides binary and quantitative results only. Generally, there are 3 types of Humaneval evaluation: pass@1, pass@10 and pass@100. They are different in the number of "chances" given to the tested model to generate the right answer to the problem. 
    \item MultiPL-E\footnote{https://huggingface.co/datasets/nuprl/MultiPL-E}: Based on the premise of HumanEval, MultiPL-E takes this benchmark and translates its results to numerous programming languages like C++, Rust, Go, Java and more. With the same ranking structure as HumanEval, this tool also provides a quantitative binary evaluation. 
    \item MBPP\footnote{https://huggingface.co/datasets/mbpp}: Consisting of approximately 1000 programming issues sourced from Python programmers, this benchmark is geared towards beginners. It offers a description of tasks, corresponding code solutions, and three automatic test cases. Its focus is on programming fundamentals and the application of standard library functions. 
    \item HumanEval Pack: This innovative evaluation method by BigCode's\footnote{https://huggingface.co/bigcode} team brings a fresh perspective to the assessment of Coding LLMs. It expands the HumanEval by engaging three different stages: Fix, Explain, and Synthesize. The “Fix” stage evaluates the model’s ability to rectify code functions containing subtle bugs, the “Explain” stage assesses the model's capacity to generate clear code explanations, while the “Synthesize" stage gauges how effectively the model synthesizes code given a natural language instruction.
\end{enumerate}

\section{Methods}
\subsection{Dataset Overview}
Overall, for the generation of Safurai-001 (starting from StarCoder 15B [\cite{starcoder}]) we used a dataset of 200,000 Q\&A examples.

As we have seen from the publications of WizardCoder [\cite{wizardcoder}] and Phi-1 [\cite{phi-1}], data quality is essential for the generation of a performing LLM coding. For this we have used the latest data augmentation and prompting engineering techniques to generate the datasets.
Furthermore, we involved some datasets and data related to basic logical and algebraic reasoning, in order to boost the comprehension StarCoder abilities.

\subsection{Initial Dataset Sources}

These are our proprietary datasets that we selected for Safurai-001 training:
\begin{itemize}
\item Safurai Code Dataset (163k)
\item Logic Q\&A Dataset (22k)
\item Math Q\&A Dataset (15k)
\end{itemize}

\subsection{Data Transformation}
We employed an additional LLM to enhance the educational potential present within the model. By incorporating both a problem and its solution, we prompted the model to elucidate the reasoning process leading to the solution. 

Our experimentation with various techniques led to the creation of a diverse dataset. The following are some of the methods we harnessed to augment the educational value:

Trasformation techniques used for our initial datasets:
\begin{itemize}
  \item Chain of thoughts reasoning
  \item Tree of thoughts reasoning
  \item Show potential errors
  \item Focus on edge cases and explain unit tests
  \item Highlight question requests in a more objective manner
  \item Coding lesson related to the topic
  \item Teaching the response
\end{itemize}

\subsubsection{Data transformation prompt experiments examples:}

\textbf{ToT Code Instructor}
\begin{lstlisting}[language=Python, 
frame=single,
xleftmargin=1cm, 
xrightmargin=1cm]
"As part of an exercise in improving AI code explanations, your task is as follows:\n"
f"Question: \n\n{row['instruction']}\n\n" 
f"Existing Answer: \n\n{row['output']}\n\n" 
"The given answer, though technically correct, doesn't offer insights into the underlying thought process.\n" 
"Your mission: devise a comprehensive step-by-step plan leading to the answer. \n" 
"This should include plain language explanations and corresponding code, neatly presented in markdown. " 
"Your answer will serve as a more informative substitute for the initial one. Strive for simplicity and human-like communication.\n\n" 
"But there's a twist: envisage a collaboration between three experts, each adding a piece to the puzzle. " 
"After contributing a step, they discuss it with the group before proceeding. " 
"If an expert determines their step is incorrect, they step away from the task. " 
"The exercise concludes when a comprehensive correct answer has been achieved, or all experts have withdrawn."
\end{lstlisting}

\textbf{CoT Code Instructor}
\begin{lstlisting}[language=Python, 
frame=single,
xleftmargin=1cm, 
xrightmargin=1cm]
"I'm training a code-writing AI and I need your help. \n" 
f"Here's a sample question: \n\n{row['instruction']}\n\n" 
f"And here's an answer: \n\n{row['output']}\n\n" 
"The given answer is too basic and doesn't explain the steps taken to arrive at it.\n" 
"Could you help create a step-by-step plan to reach this answer? \n" 
"Each step should be simple and understandable. \n" 
"Your answer should include this plan and the actual code in markdown in one block." 
"Your answer will replace the one I've shown you. It should sound human!" 
"Make sure not to cut off words or sentences midway."
\end{lstlisting}

\textbf{Teacher Code Instructor}
\begin{lstlisting}[language=Python, 
frame=single,
xleftmargin=1cm, 
xrightmargin=1cm]
"Imagine you are a programming expert tasked with providing clear and formal programming assistance. " 
f"You are presented with this problem: \n\n{row['instruction']}\n\n" 
"You have two primary goals: " 
"1) explain the process to solve the problem step by step. Explain the process to solve the problem step by step in a conversational manner, with a few bullet points." 
"2) include specific examples of common errors that should be avoided, accompanied by code snippets illustrating these mistakes. Tag these code snippets as 'Error Example'. " 
"When providing the correct solution, ensure there are comments in the code to enhance its comprehensibility, addressing crucial points and possible mistakes. "
\end{lstlisting}

\textbf{CoT Logic Instructor}
\begin{lstlisting}[language=Python, 
frame=single,
xleftmargin=1cm, 
xrightmargin=1cm]
"I have a dataset with questions and responses about logical problems. \n" 
f"This is one logical problem: \n\n{row['instruction']}\n\n" 
f"This is the provided solution to the problem: \n\n{row['output']}\n\n" 
"The provided solution is too simple and doesn't explain the process to get it.\n" 
"Can you please provide a detailed reasoning to get the solution? \n"
\end{lstlisting}

\subsection{Safurai-001 Datasets used for finetuning:}
\begin{itemize}
\item Safurai Code Instructor (16k) (filtered examples from initial Safurai Code Dataset, not transformed)
\item Logic Q\&A Dataset (22k)
\item Teacher Code Instructor (70k)
\item Math Q\&A Dataset (15k)
\item Teacher Code Instructor with Potential Errors (21k)
\item ToT Code Instructor (30k)
\item CoT Code Instructor (26k)
\end{itemize}

\begin{figure}[h]
\centering
\includegraphics[width=1\textwidth]{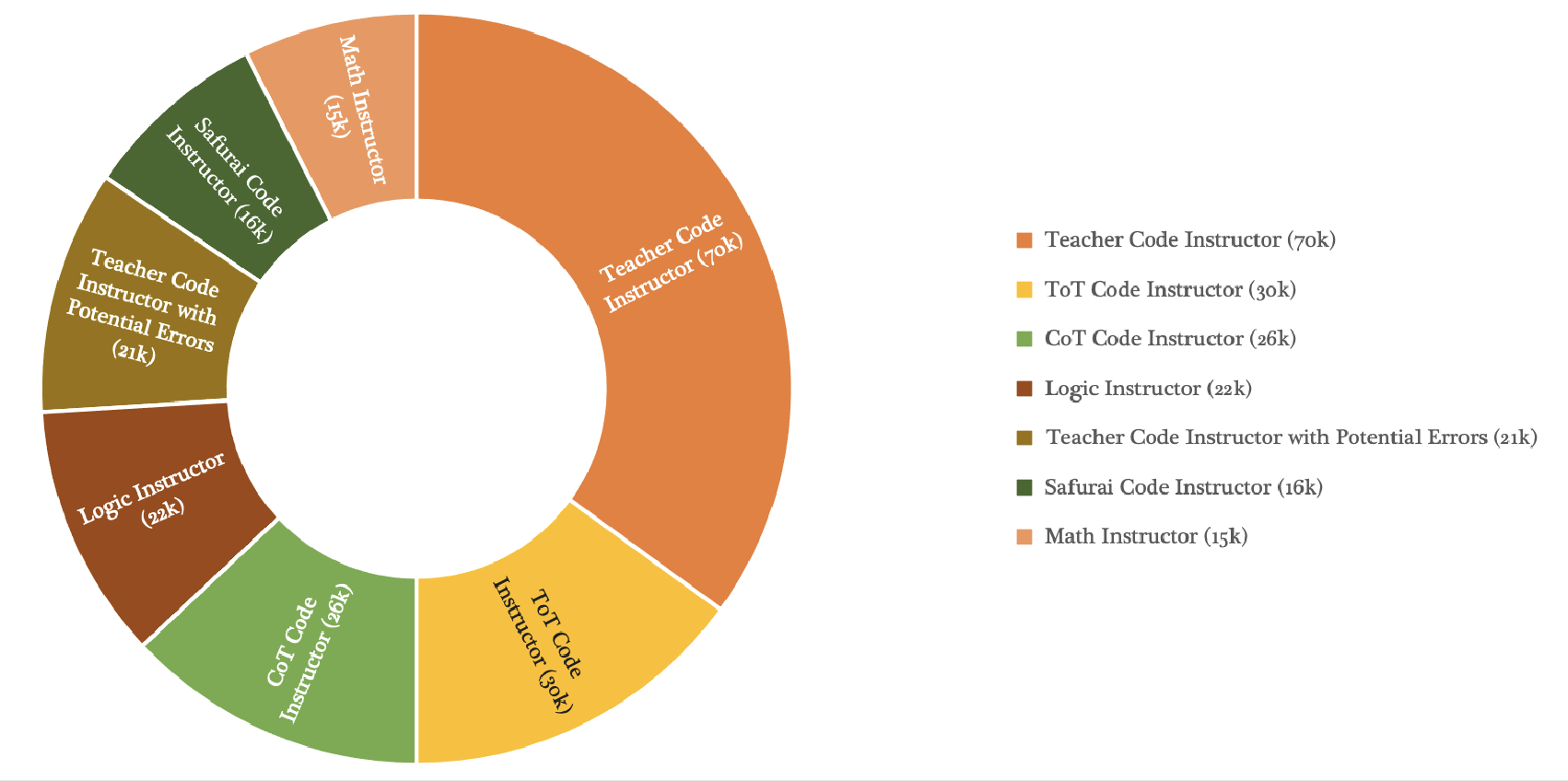}  
\caption{Safurai-001 Datasets Compartition}
\label{fig:my_label}
\end{figure}

\subsection{Fine-Tuning Methods}
The training process for the StarCoder model was carried out on eight 100 A100 80GB graphics cards, with the full network trained for 10 hours. This process was implemented using the DeepSpeed ZeRO-3 framework. In order to fine-tune the StarCoder models, we leveraged the GitHub repository of Big Code: \href{https://github.com/bigcode-project/starcoder/tree/main/chat}{https://github.com/bigcode-project/starcoder/tree/main/chat}

Given that the focus was on instruction tuning, the data was wrapped in a dialogue format. An example of this is illustrated below:

\begin{lstlisting}[language=Python, 
frame=single,
xleftmargin=1cm, 
xrightmargin=1cm]
{
    "messages": [
        {
            "content": "Is it possible to imagine a society without law?", 
            "role": "user"},
        {
            "content": "It is difficult to imagine a society that is able to be maintained without any semblance of Law.",
            "role": "assistant",
        },
        {
            "content": "It seems like you consider the absence of law equal to the absence of anything that could guide the behaviour of the individual.",
            "role": "user",
        },
        {
            "content": "You are correct that there are other factors that can guide behavior in a society and play a role in shaping individuals' behavior and interactions with each other. However, even in societies where these factors are present, laws still serve an important role in maintaining social order and resolving conflicts.",
            "role": "assistant",
        }
    ]
}
\end{lstlisting}

We also wrapped the dialogue with special tokens. Here's an example of how it looks:

\begin{lstlisting}[language=Python, 
frame=single,
xleftmargin=1cm, 
xrightmargin=1cm]
<|system|>
Below is a dialogue between a human and AI assistant called StarChat.
<|end|>
<|user|>
Is it possible to imagine a society without law?<|end|>
<|assistant|>
It is difficult to imagine ...<|end|>
<|user|>
It seems like you ...<|end|>
<|assistant|>
You are correct ...<|end|>
<|user|>
Yeah, but laws are complicated ...<|end|>
\end{lstlisting}

The training process involved setting up several hyperparameters. The hyperparameters settings for the training are detailed as follows:
\begin{itemize}
    \item Batch size: 512
    \item Learning rate: 2e-5
    \item Epochs: 3
    \item Max length: 2048
    \item Warmup step: 30
    \item Learning rate (LR) scheduler: cosine
\end{itemize}

\par
\subsection{Evaluation}
Deepening our grasp of the capabilities and scope of LLM models is substantial to refining their application in the real world. However, we found the currently available evaluation methods such as HumanEval\footnote{https://huggingface.co/datasets/openai\_humaneval} to be limited in their ability to provide a comprehensive analysis of these models' abilities. This led to the invention of the GPT4-based MultiParameters Evaluation method, a qualitative alternative designed to provide a more nuanced understanding of the performance of coding LLMs. 

These new qualitative criteria enable us to explore more use-cases outside the conventional binary pass-fail result of the existing quantitative methods, thus providing a more detailed narrative that identifies the unique strengths (or weaknesses) of each model.
HumanEval, MBPP\footnote{https://huggingface.co/datasets/mbpp} and MultiPL-E\footnote{https://huggingface.co/datasets/nuprl/MultiPL-E}. Most of them lean towards a quantitative rather than a qualitative evaluation, leaving out crucial aspects of the models' capabilities. As such, we justify the innovation and necessity of our GPT4-based MultiParameters Evaluation method in addressing this gap.

\subsection{New experiments on the evaluation (GPT4-Based MultiParameters Evaluation)}
Seeking to explore the qualitative aspects of our LLM model Safurai, we experimented with a new evaluation approach based on GPT-4 [\cite{gpt4}]. 
\subsubsection{GPT4-based Analysis}
This method involved assessing 20 (GPT-4 HE-20) and 40 (GPT-4 HE-40) answers derived from each of the models being compared, obtained through the HumanEval dataset. GPT-4 was used to determine the performance of models: Safurai, Claude\footnote{https://www.anthropic.com/index/introducing-claude}, WizardCoder [\cite{wizardcoder}], ChatGPT\footnote{https://openai.com/blog/chatgpt}, and Starchat Alpha Prompted\footnote{https://huggingface.co/HuggingFaceH4/starchat-alpha}. We combined both the problem and the five responses into a single GPT-4 prompt plus the specific tests of the problem, which was asked to rate each response on a scale of 0 to 100 – the best possible score. Moreover, to deepen our understanding, we asked GPT-4 to provide a concise description detailing the reasoning behind its ratings. 
This is the GPT-4 prompt:
\textit{I asked this to 4 different AI models: [problem]
This is the first model answer: [answer]
This is the second model answer: [answer]
This is the third model answer: [answer]
This is the fourth model answer: [answer]
These are the tests for the code solution of the problem: [tests]
Please rate each answer from 0 to 100 (best answer possible). Consider whether the code fully solves the problem, if it handles all edge cases, and if it contains all necessary functionalities. Also, provide a short explanation for each rating.}
This way, in addition to quantifying performance, our evaluation strategy reveals valuable insights into each model's strengths and weaknesses. (Table 2)

\begin{center}
    Table 2: Results of GPT-4 HumanEval-20 and GPT-4 HumanEval-40
\end{center}
\renewcommand{\arraystretch}{1.8} 
\setlength{\tabcolsep}{5pt} 

\begin{tabularx}{\textwidth} { 
  >{\centering\arraybackslash}X
  >{\raggedright\arraybackslash}X 
  >{\centering\arraybackslash}X 
  >{\centering\arraybackslash}X}  
 \hline
  \textbf{Date} & \textbf{Model} & \textbf{GPT-4 HumanEval-20} & \textbf{GPT-4 HumanEval-40} \\
 \hline
 \multicolumn{4}{c}{\textbf{Closed source models}} \\
 \hline
 2022 Nov & GPT3.5-turbo  & 81,5\% & 80,875\% \\  
 2023 March & Claude  & 75\% & 78,7\% \\  
 \hline
 \multicolumn{4}{c}{\textbf{Open source models}} \\ 
 \hline
 2023 May & Starchat-Alpha prompted  & 64,3\% & 62,4\% \\  
 2023 June & WizardCoder  & 74,4\% & 74,7\% \\  
 \hline
 2023 June & \textbf{Safurai-001}  & \textbf{85\%} {(\color{red}+3.5\%}) & \textbf{84,875\%} {(\color{red}+4\%}) \\  
\hline
\end{tabularx}

The experiments detailed above provided a holistic process for comparative model evaluation. By evaluating 20 (GPT-4 HE-20) and 40 (GPT-4 HE-40) responses from each compared model using the HumanEval dataset, we generated valuable quantitative data and underlying qualitative insights on model performance.

However, we recognized that the comprehensive ratings provided by GPT-4, while integral to the evaluation process, cannot fully capture the nuanced specificities inherent in each model. Comprehensive ratings bootstrap a model's ability to resolve a problem and generate correct code, but they fall short in illuminating aspects such as efficiency, readability, best coding practices, and relevance to problem. These key dimensions, though less evident, are equally vital to a model's utility and impact in real-world software development scenarios.

To alleviate these shortcomings and provide a more detailed, multidimensional, and nuanced appraisal of the models' functionalities, we introduced a four-parameter rating system.

\subsubsection{GPT4-based MultiParameters Evaluation Benchmark}
To understand even more about the model responses, we created a Multi-Parametric GPT4-based Evaluation system. The singular GPT-4 prompt, containing both the problem and the four respective solutions, was not only rated generally but was also dissected based on four distinct parameters. These were:
\begin{enumerate}
    \item Code Correctness and Completeness: This involved gauging whether the code runs without errors and if it fully solves the problem, considering all potential edge cases.
    \item Efficiency: This measurement determined the optimization level of the code. It scrutinized whether the code utilizes resources capably, and whether it scales efficiently as input size increases.
    \item Readability and Best Practices: This criterion evaluated the clarity of the written code, whether it's easily comprehensible, and if it conforms to established coding conventions and best practices.
    \item Relevance to Problem (On-point Answer): This parameter evaluated how directly the code solves the given problem, assessing whether the solution implemented is efficacious and appropriate.
\end{enumerate}

These are the GPT-4 prompts used for each parameter:
\begin{enumerate}
    \item \textit{I asked this to 4 different AI models: [problem]
This is the first model answer: [answer]
This is the second model answer: [answer]
This is the third model answer: [answer]
This is the fourth model answer: [answer]
These are the tests for the code solution of the problem: [tests]
Please rate each answer from 0 to 100 (best answer possible) based on Code Completeness. Consider whether the code fully solves the problem, if it handles all edge cases, and if it contains all necessary functionalities. Also, provide a short explanation for each rating.}
    \item \textit{I asked this to 4 different AI models: [problem]
This is the first model answer: [answer]
This is the second model answer: [answer]
This is the third model answer: [answer]
This is the fourth model answer: [answer]
These are the tests for the code solution of the problem: [tests]
Please rate each answer from 0 to 100 (best answer possible) on Efficiency. This entails considering how well-optimized the code is, how frugally it uses system resources, and its scalability or robustness for larger inputs. Consider both its time complexity (ability to perform tasks quickly) and space complexity (how much memory the program uses). Also, provide a short explanation for each rating.}
    \item \textit{I asked this to 4 different AI models: [problem]
This is the first model answer: [answer]
This is the second model answer: [answer]
This is the third model answer: [answer]
This is the fourth model answer: [answer]
These are the tests for the code solution of the problem: [tests]
Please rate each answer from 0 to 100 (best answer possible) based on its Helpfulness and Educational Value. Consider whether the answer provides clear explanations, whether it's easy to follow and understand, whether it teaches you something valuable about the problem or the coding concepts involved, and whether it gives you new insights that could help you in future similar problems. Also, provide a short explanation for each rating.}
    \item \textit{I asked this to 4 different AI models: [problem]
This is the first model answer: [answer]
This is the second model answer: [answer]
This is the third model answer: [answer]
This is the fourth model answer: [answer]
These are the tests for the code solution of the problem: [tests]
Please rate each answer from 0 to 100 (best answer possible) based on its Relevance to Problem (On-point answer). Consider how directly the code answers the problem, if it provides an adequate and appropriate solution, and whether it implements the requirements stated in the problem. Also, provide a short explanation for each rating.}
\end{enumerate}

\subsubsection{Why GPT-4?}
Choosing GPT-4 as the foundation for our novel evaluation method was a clear and strategic decision. At the current state-of-the-art, GPT-4 showcases the most superior performance amongst all existing coding models. 

In addition to its exemplary robustness and efficiency, GPT-4 boasts an extensive range of applications, making it an ideal model for evaluating less potent ones in diverse facets. This ability allows the evaluation process to be as comprehensive as possible, scrutinizing every model under multiple parameters and therefore obtaining an all-encompassing performance profile.

However, it's worth noting that there are inherent limitations associated with GPT-4. These limitations, reflected as weaknesses in our newly proposed evaluation system, include potential biases carried over from GPT-4 behaviours. Nevertheless, as long as no other model matches or surpasses GPT-4's aptitude, these weaknesses remain relatively negligible. We reckon that at this stage, GPT-4 presents the most accurate measure for performance indicators considered in our GPT4-based MultiParameters Evaluation methodology.

Thus, until a more advanced model than GPT-4 enters the arena, our focus remains on deriving maximum benefit from this superior model, optimizing its use in revealing the strengths and areas of improvement of each evaluated coding model.

\section{Experimental Result}
\subsection{Model Performances}
\textbf{Example of Multi-Turn conversation with Safurai-001 }

\begin{tcolorbox}
    \textbf{USER}: Hello Safu!
\end{tcolorbox}

\begin{tcolorbox}
    \textbf{Safurai-001}: Hello! How can I assist you today?
\end{tcolorbox}

\begin{tcolorbox}
    \textbf{USER}: What's your level of coding in Python?
\end{tcolorbox}

\begin{tcolorbox}
    \textbf{Safurai-001}: As an AI language model, I have been created and trained to have a strong understanding of Python programming language. I can help with a wide range of topics and tasks in Python, from basic syntax and concepts to more advanced algorithms or frameworks. Feel free to ask me anything related to Python coding, and I'll do my best to assist you!
\end{tcolorbox}

\begin{tcolorbox}
    \textbf{USER}: How can I write a function in Python to find the median of a list? Please include the time and space complexity of the solution.
\end{tcolorbox}

\begin{tcolorbox}
    \textbf{Safurai-001}: The median of a list is the middle value when the list is sorted in ascending order. If the list has an odd number of elements, the median is the middle element. If the list has an even number of elements, the median is the average of the two middle elements.

To solve this problem, we can follow these steps:
\begin{enumerate}
    \item Sort the list in ascending order.
    \item Check if the length of the list is odd or even.
    \item If the length is odd, return the middle element.
    \item If the length is even, calculate the average of the two middle elements and return it.
\end{enumerate}

Here is the implementation of the median function:
\begin{lstlisting}[ 
frame=single,
xleftmargin=2cm, 
xrightmargin=2cm]
def median(l: list):
    l.sort()  # Sort the list in ascending order
    length = len(l)
    if length % 2 == 1:  # Check if the length is odd
        return l[length // 2]  # Return the middle element
    else:
        mid1 = l[length // 2]  # First middle element
        mid2 = l[length // 2 - 1]  # Second middle element
        return (mid1 + mid2) / 2  # Return the average of the two middle elements
\end{lstlisting}
Test cases:
\begin{lstlisting}[ 
frame=single,
xleftmargin=2cm, 
xrightmargin=2cm]
print(median([3, 1, 2, 4, 5]))  # Output: 3
print(median([-10, 4, 6, 1000, 10, 20]))  # Output: 15.0
\end{lstlisting}
The time complexity of this solution is O(n log n), where n is the length of the input list. This is because the sorting operation takes O(n log n) time. The space complexity is O(1) as we are not using any additional data structures.
\end{tcolorbox}

\subsection{Transformed Dataset Examples}

\textbf{Original Input}
\begin{tcolorbox}
    Write a JavaScript code to loop over all elements in the given array. Input: numbersArray = [45, 6, 23, 12, 35]
\end{tcolorbox}

\textbf{Original Output}
\begin{tcolorbox}
    for(let i = 0; i < numbersArray.length; i++) { console.log(numbersArray[i]); }
\end{tcolorbox}

\textbf{Augmented Output}
\begin{tcolorbox}
Sure, I'd be happy to help! 1. Start by declaring the array variable:
\begin{lstlisting}[ 
frame=single,
xleftmargin=2cm, 
xrightmargin=2cm]
const numbersArray = [45, 6, 23, 12, 35]
\end{lstlisting}
2. We want to loop over all elements in the array, so we need to use a for loop. 
3. Declare our index variable i and set it to 0
4. Define the loop condition; we want to continue looping as long as the index variable is less than the length of the array. 5. Use the console.log() method to print out the value of the current element in the array. 6. Finally, increment the index variable by 1 to move onto the next element in the array. Here's the final code:
\begin{lstlisting}[ 
frame=single,
xleftmargin=2cm, 
xrightmargin=2cm]
const numbersArray = [45, 6, 23, 12, 35]; for (let i = 0; i < numbersArray.length; i++) { console.log(numbersArray[i]); }
\end{lstlisting}
This code will loop over all elements in the array and print out their values to the console.
\end{tcolorbox}

\subsection{Evaluation Results}
\subsubsection{HumanEval}
HumanEval\footnote{https://github.com/openai/human-eval}, established by OpenAI, has been the go-to benchmark in evaluating coding AI models. It focuses on the model's code-generation capacity based on precise requests and provides standard solutions and tests. MultiPL-E\footnote{https://huggingface.co/datasets/nuprl/MultiPL-E/viewer/humaneval-rs/test?row=0} extends the boundaries of the current benchmarks by translating them to new languages and thereby becoming a massive multi-language benchmarking platform. It is continuously expanding in terms of number of programming languages, providing a useful comparison point for models like Safurai.

Our model, Safurai001, achieved a pass@1 score of 50.61\% on the HumanEval benchmark with n=20. (Table 1)

\begin{table}[h]
\centering
\caption{Results of pass@1(\%) on HumanEval}
\label{tab:performance}
\renewcommand{\arraystretch}{1.5}
\scalebox{0.8}{
\begin{tabularx}{\textwidth}{ 
  >{\hsize=1.5\hsize\centering\arraybackslash}X
  >{\hsize=.5\hsize\raggedright\arraybackslash}X 
  >{\hsize=.5\hsize\centering\arraybackslash}X}  
 \hline
  \textbf{Model} & \textbf{Size} & \textbf{HumanEval} \\
 \hline
 \multicolumn{3}{c}{\textbf{Closed source models}} \\
 \hline
Alphacode Li et al. [2022] & 540B & 26.2 \\
Codex Chen et al. [2021] & 12B & 28.8 \\
Code-Cushman-001 OpenAI [2022] & - & 33.5 \\
Code-Davinci-002 OpenAI [2022] & - & 47.0 \\
GPT-3.5 OpenAI [2023] & - & 48.1 \\
GPT-3.5 Luo et al. [2023] & - & 68.9 \\
GPT-4 OpenAI [2023] & - & 67.0 \\
GPT-4 Bubeck et al. [2023] & - & 82.0 \\
\hline
\multicolumn{3}{c}{\textbf{Open source models}} \\ 
\hline
LLaMa Touvron et al. [2023] & 65B & 23.7 \\
CodeT5+ Wang et al. [2023] & 16B & 30.9 \\
StarCoder Li et al. [2023] & 15B & 33.6 \\
WizardCoder Luo et al. [2023] & 15B & 57.3 \\
\hline
\textbf{Safurai001 [2023]} & \textbf{15B} & \textbf{50.61} \\
\hline
\end{tabularx}
}
\end{table}
 
However, the adoption of only these standards limits our analysis to quantitative metrics, thereby losing some critical flavors of the models.

\subsubsection{New Qualitative Evaluation Benchmark}
We tested the models with the 40 selected problems of HumanEval, already used in GPT4-based Analysis.
The GPT4-based MultiParameters Evaluation method elucidates areas for optimization, explains why a specific response is superior, and significantly comprehends the specific code-generation abilities of each model; thus providing a detailed qualitative metric.

We found that this method reveals a plethora of valuable insights into each model's strengths and weaknesses, enabling the development of targeted strategies for enhancement. (Table 3)

\begin{table}[h]
\centering
\caption{Results of GPT4-based MultiParameters HumanEval}
\label{tab:performance2}
\renewcommand{\arraystretch}{1.5}
\scalebox{1}{
\begin{tabularx}{\textwidth}{ 
  >{\hsize=1\hsize\centering\arraybackslash}X
  >{\hsize=1\hsize\centering\arraybackslash}X 
  >{\hsize=1\hsize\centering\arraybackslash}X 
  >{\hsize=1\hsize\centering\arraybackslash}X
  >{\hsize=1\hsize\centering\arraybackslash}X  
  >{\hsize=1\hsize\centering\arraybackslash}X}
 \hline
  \textbf{Date} & \textbf{Model} & \textbf{Code Correctness} & \textbf{Code Efficiency} & \textbf{Code Readability} & \textbf{Question Relevance} \\
\hline
2022 Nov & GPT-3.5-turbo & 81.53\% & 80.33\% & 84.30\% & 82.25\% \\
2023 March & GPT-4 & 89.50\% & 89.38\% & 84.10\% & 90.93\% \\
2023 June & WizardCoder & 60.7\% & 68.25\% & 67.1\% & 67.88\% \\
\hline
\textbf{2023 July} & \textbf{Safurai-001} & \textbf{74.25\%} & \textbf{75.45\%} & \textbf{85.88\%(+1.58\%)} & \textbf{82.00\%} \\
\hline
\end{tabularx}
}
\end{table}

We put our proposed GPT4-based MultiParameters Evaluation method to the test, using the same 40 selected problems from HumanEval which had been previously used in our GPT4-based Analysis. The results obtained were intriguing, enlightening and informative, revealing areas of optimization and superiority in specific responses and highlighting the need to explore code-generation abilities at a profound level. The qualitative data provided by this method was a treasure trove of information, reaching depths previous evaluation methods did not venture.

Interestingly, this assessment unveiled nuances in model performance that were not entirely predictive of functionality during actual deployment. For instance, despite WizardCoder [\cite{wizardcoder}] achieving higher scores in the HumanEval evaluation, it was observed that real-world day-to-day usage, especially for developers, was not as smooth. The model’s conversational abilities seemed to be somewhat lacking, making it hard to interact effectively with it. This was reflected in its score of 67.1 in the Code Readability category, a stark contrast with Safurai001's impressive score of 85.88. 

In shadowing the performance of conventional quantitative benchmarks like HumanEval and MultiPL-E, we developed a new qualitative evaluation method: the GPT4-based MultiParameters Evaluation. This unprecedented approach provided a broader perspective of the nuances and intricacies of LLM models, broadening the spectrum of their functionality and applications.

Models like Phi1 [\cite{phi-1}], developed by Microsoft Researchers, StarCoder [\cite{starcoder}], and WizardCoder [\cite{wizardcoder}], are mainly evaluated using conventional methods. While efficient, these methods lack the ability to provide an exhaustive understanding of the model’s capabilities, thus justifying the necessity for the development of our new evaluation method.

The GPT4-based MultiParameters Evaluation method breaks new ground in the LLM models evaluation field, enabling researchers to delve deeper into the functionality of these models and dramatically enhancing their potential for improvement.

\subsubsection{Limits of GPT4-based MultiParameter Evaluation Benchmark}

\begin{itemize}
    \item \textbf{The GPT4-based MultiParameters Evaluation method can only evaluate up to a certain limit}: It is benchmarked at the GPT4 level of performance. This implies that any model exceeding the performance of GPT4 may not be reliably evaluated or accurately assessed. This method, therefore, does not cater to the fast-paced advancements and continuous improvements in the AI development landscape. 
    \item \textbf{Variance in responses:} Another constraint associated with the GPT4-based MultiParameters Evaluation method is the variation in GPT4's response. It does not always provide consistent responses due to various reasons such as the difference in nature and complexity of the problem. This inconsistency presents challenges in providing a robust evaluation of the Coding AI models. 
    \item \textbf{Time-intensive:} The GPT4-based MultiParameters Evaluation is a relatively time-consuming process. Due to the complex and multifaceted nature of this evaluation method, it requires substantial time and resource compared to conventional methods. This could limit its scalability and practical application in rapidly iterating development cycles. 
    \item \textbf{Qualitative nature:} The GPT4-based MultiParameters Evaluation method mainly provides a qualitative evaluation. It may not effectively handle quantitative data or metrics, and comparing those could be challenging. This could present limitations when models' evaluation requires numerical comparisons or precise quantifiable results. 
\end{itemize}

\subsubsection{Example of GPT4-based MultiParameter Evaluation response}
This is an example of GPT4-based MultiParameter Evaluation response. The problem is the Number 0 of HumanEval Benchmark. About the answers, the first model is Safurai-001, the second ChatGPT, the third WizardCoder, and the fourth GPT4.

\textbf{First Parameter: Code Correctness}

\begin{tcolorbox}
    First Model Answer:
Rating: 90
Explanation: The code provided by the first model is thorough, complete and well-explained. It correctly implements the function to solve the problem. It handles the basic functionality of checking if any two numbers in the list are closer than the threshold. However, it does not explicitly handle the edge case where the list of numbers is empty.

Second Model Answer:
Rating: 90
Explanation: The second model provides a complete solution that is functional and concise. However, it does not handle edge case where the list could be empty or include only one element.

Third Model Answer:
Rating: 80
Explanation: The third model provides a functional solution to the problem, but it lacks in explanation and handling edge case when the list is empty or includes only one element.

Fourth Model Answer:
Rating: 100
Explanation: The fourth answer provides two complete solutions, one with a time complexity of $O(n^2)$ and another efficient one with $O(n log n)$. It correctly handles all scenarios and edge cases and provides a good balance between explanation, completeness and efficiency.
\end{tcolorbox}

\textbf{Second Parameter: Code Efficiency}
\begin{tcolorbox}
Answer 1: 75/100
This answer provides a brute-force solution by comparing each element with every other element in the list. Although clear and correct, this solution has a time complexity of $O(n^2)$, which means it may not be efficient for large lists of numbers, as it doesn't scale well with increasing input size.

Answer 2: 75/100
The solution given in Answer 2 is almost identical to the one in Answer 1, having a time complexity of $O(n^2)$ with a pairwise comparison. This lacks efficiency for large inputs, and could benefit from using more optimized approaches, such as sorting the list first.

Answer 3: 75/100
Much like the first two answers, answer 3 also uses a brute-force approach in the solution. As it compares all pairs of numbers, it has a time complexity of $O(n^2)$, meaning it won't be efficient for larger lists.

Answer 4: 90/100
Answer 4 is the best of the lot in terms of efficiency. The answer actually lists two different approaches: a brute-force approach mentioned in the first three answers, and a more efficient approach that first sorts the list before comparing adjacent numbers. The sorted approach has a time complexity of $O(n log n)$, which scales better with a growing list of numbers than the $O(n^2)$ complexity of the brute-force methods. The only point deduction is due to not explicitly explaining this crucial difference in complexity between the two methods.
\end{tcolorbox}
\newpage
\textbf{Third Parameter: Code Readability}
\begin{tcolorbox}
First model's answer:
Rating: 90
Explanation: The first model provides a detailed explanation of the problem, the method used to solve it, and even breaks down the entire code, making it easy to understand for beginners too. The example provided at the end adds an extra value to the answer as well. Yet, it still fails to explain or consider any efficiency measures or alternate ways to solve the problem.

Second model's answer:
Rating: 70
Explanation: The second model also correctly interprets the problem and comes up with a valid approach. However, it's less detailed than the first model, doesn't provide a clear breakdown of the code, and offers no example of how to use the function.

Third model's answer:
Rating: 35
Explanation: The third model just provides the code needed to solve the problem without any explanations or breakdown of the code. Its answer is correct but not very helpful or educational, especially for someone trying to learn.

Fourth model's answer:
Rating: 95
Explanation: The fourth model does a great job of explaining the problem and also introduces two different approaches to solve it, one being more efficient than the other. It not only offers insight into a more effective solution but also explains why it's more efficient. But despite these advantages, the model did not break down the code or provide examples of usage.
\end{tcolorbox}

\textbf{Fourth Parameter: Question Relevance}
\begin{tcolorbox}
Rating for the first model: 90.
The answer from the first model is quite thorough and directly addresses the problem. The explained code solves the problem as required and it perfectly implements the requirement. However, it does not consider the efficiency of the solution.

Rating for the second model: 85.
The second model also provides an adequate answer, explaining the solution fairly well and implementing the described function correctly. However, much like the first model, this model does not consider any performance or efficiency aspects in its solution.

Rating for the third model: 80.
This AI model provided a correct and direct solution, but with less explanation about the approach or notes on performance or usage. Therefore, the rating is lower compared to the first and second models.

Rating for the fourth model: 100.
The fourth model provided two solutions: the first one similar to the previous models and the second one that considers performance improvements by sorting the list first. This model shows a greater understanding of real-world requirements by considering efficiency, thus making it the best answer.
\end{tcolorbox}

\subsubsection{Comparison of GPT4-based MultiParameters and HumanEval benchmarks}

\begin{center}
    \textbf{Problem}
\end{center}

\begin{lstlisting}[language=Python, 
frame=single,
xleftmargin=2cm, 
xrightmargin=2cm]
from typing import List

def has_close_elements(numbers: List[float], threshold: float) -> bool:
    """
    Check if in given list of numbers, are any two numbers closer to each 
    other than given threshold.

    >>> has_close_elements([1.0, 2.0, 3.0], 0.5)
    False
    >>> has_close_elements([1.0, 2.8, 3.0, 4.0, 5.0, 2.0], 0.3)
    True
    """
\end{lstlisting}

\begin{center}
    \textbf{Safurai-001 Response}
\end{center}

First, we need to understand what the problem is asking us to do. We have a list of numbers and we need to check if any two of those numbers are closer than a certain threshold. 

To solve this problem, we can use a loop to compare each number in the list to every other number in the list. If the difference between the two numbers is less than or equal to the threshold, we know that they are closer than the threshold. 

Here is the code to solve this problem:
\begin{lstlisting}[language=Python, 
frame=single,
xleftmargin=2cm, 
xrightmargin=2cm]
# Define a function that takes in a list of numbers and a threshold
def has_close_elements(numbers: List[float], threshold: float) -> bool:
    # Loop through each number in the list
    for i in range(len(numbers)):
        # Loop through every other number in the list
        for j in range(i+1, len(numbers)):
            # Check if the difference between the two numbers is less than or equal to the threshold
            if abs(numbers[i] - numbers[j]) <= threshold:
                # If it is, return True
                return True
    # If we have checked every pair of numbers and none of them are closer than the threshold, return False
    return False
\end{lstlisting}
Let's break down the code:
\begin{itemize}
    \item `def has\_close\_elements(numbers: List[float], threshold: float) -\texttt{>} bool:`: This line defines a function called `has\_close\_elements` that takes in a list of numbers (`numbers`) and a threshold (`threshold`) and returns a boolean value (either `True` or `False`).
    \item `for i in range(len(numbers)):`: This line starts a loop that goes through each number in the list.
    \item `for j in range(i+1, len(numbers)):`: This line starts a nested loop that goes through every other number in the list (starting from the next number after `i`).
    \item `if abs(numbers[i] - numbers[j]) \texttt{<}= threshold:`: This line checks if the difference between the two numbers is less than or equal to the threshold.
    \item `return True`: If the two numbers are closer than the threshold, we return `True`.
    \item `return False`: If we have checked every pair of numbers and none of them are closer than the threshold, we return `False`.
\end{itemize}

To use this function, we can call it with a list of numbers and a threshold:
\begin{lstlisting}[language=Python, 
frame=single,
xleftmargin=2cm, 
xrightmargin=2cm]
# Call the function with a list of numbers and a threshold
result = has_close_elements([1.0, 2.8, 3.0, 4.0, 5.0, 2.0], 0.3)

# Print the result
print(result) # Output: True
\end{lstlisting}
In this example, the list of numbers `[1.0, 2.8, 3.0, 4.0, 5.0, 2.0]` has two numbers that are closer than the threshold of `0.3`. The function returns `True` because at least one pair of numbers is closer than the threshold.

\begin{center}
    \textbf{ChatGPT Response}
\end{center}

The code you provided defines a function called has1\_close\_elements that takes two arguments: numbers, which is a list of float numbers, and threshold, which is a float representing the maximum allowed difference between two numbers.
The function checks if there are any two numbers in the numbers list that are closer to each other than the given threshold. It returns True if such a pair exists, and False otherwise.
Here's an implementation of the has\_close\_elements function based on the code you provided:
\begin{lstlisting}[language=Python, 
frame=single,
xleftmargin=2cm, 
xrightmargin=2cm]
from typing import List

def has_close_elements(numbers: List[float], threshold: float) -> bool:
    for i in range(len(numbers)):
        for j in range(i + 1, len(numbers)):
            if abs(numbers[i] - numbers[j]) < threshold:
                return True
    return False
\end{lstlisting}
This implementation uses nested loops to compare each pair of numbers in the numbers list. The abs() function is used to calculate the absolute difference between the two numbers, and if this difference is less than the threshold, the function returns True. If no such pair is found after examining all combinations, the function returns False.
You can use this function to check if there are any close elements in a list of numbers as demonstrated in the examples you provided.
\begin{center}
    \textbf{HumanEval Benchmark for this responses}
\end{center}

For HumanEval tests, both of Safurai-001 and ChatGPT answers are correct in the same way.
\begin{center}
    \textbf{GPT4-Based MultiParameters Evaluation Benchmark for this responses}
\end{center}

The first model is Safurai-001, the second one is ChatGPT.

\textbf{Code Correctness}
\par
First model answer rating: 95
The code given here is complete and handles the problem well by checking every pair of numbers for a difference less than or equal to the given threshold. It gives a full explanation of the logic and running of the code, and creates a clear distinction of cases when it should return `True` or `False`. However, the instruction could have been clearer for the function to return `True` if the difference is smaller than the threshold, not equal or smaller.

Second model answer rating: 95
This answer very similarly provides complete code that solves the problem, considering all possible pairs of numbers in the given list and checking if the difference is lesser than the provided threshold. It also uses good explanations to clarify how the code works. However, the second model doesn't make the distinction in the condition checking as clearly as the first answer. And like the first answer, it also considers equals to threshold as a valid condition.

\textbf{Code Efficiency}
\par
First model's answer rating: 90/100. The proposed algorithm takes a straightforward approach to solve the problem with a time complexity of $O(n^2)$, given n is the length of the input list. For each element i in the list, it compares it with every other element j in the list to see if the absolute value of their difference is less or equal to the threshold. Because this model only focuses on the pairwise comparison without any optimization (like sorting or reducing the search space), it might not be the most efficient solution when the input size increases, hence a minor deduction in the rating.

Second model's answer rating: 90/100. By comparison, the second model also produces a correct $O(n^2)$ solution, which uses two nested loops to compare every unique pair of values in the provided list. However, similar to the first model, it presents an algorithm that might not perform well for very long lists, and it does not provide ways to optimize or scale the operation. Therefore, it also does not get the full rating. Both performed similarly in terms of code efficiency.

\textbf{Code Readability}
\par
First model answer: 
Rating: 95
The answer elaboratively explains the problem and its solution. It clearly describes how the function works, and the step-by-step breakdown of the code is beneficial in understanding the algorithm. The example at the end, which demonstrates how to use the function, is also very helpful.

Second model answer: 
Rating: 80
This answer also gives a clear explanation of the solution and function. However, it lacks an in-depth breakdown of the code and doesn't provide an example of how to use the function, which would have been useful for demonstrating its functionality.

\textbf{Question Relevance}
\par
First model answer rating: 95/100
This response not only provides the correct implementation of the function, but also explains the solution in a detailed manner. It thoroughly covers how the function works, including explaining its various parts like two nested loops, use of abs() function and the if conditions. The example usage was an added bonus. I've docked a few points because the model does not talk about the time complexity of this solution and also does not provide the full code snippet which is importing necessary modules from the typing module.

Second model answer rating: 85/100
This response also delivers a correct code implementation with a succinct explanation. Though it explains the working of the function effectively, it's not as detailed as the first model's answer. This response is also short on providing information on usage of the function and discussing the time complexity of the solution. Importantly, it doesn't restate the question's example into code for demonstrating purposes, which could be highly informative for users who are learning or new to Python. Therefore, I've rated it slightly lower.

\begin{table}[h]
\centering
\caption{Results of GPT4-based MultiParameters HumanEval for HumanEval Problem 0}
\label{tab:performance3}
\renewcommand{\arraystretch}{2.5}
\scalebox{1}{
\begin{tabularx}{\textwidth}{ 
  >{\hsize=1\hsize\centering\arraybackslash}X 
  >{\hsize=1\hsize\centering\arraybackslash}X 
  >{\hsize=1\hsize\centering\arraybackslash}X
  >{\hsize=1\hsize\centering\arraybackslash}X  
  >{\hsize=1\hsize\centering\arraybackslash}X}
 \hline
 \hline
   & \textbf{Code Correctness} & \textbf{Code Efficiency} & \textbf{Code Readability} & \textbf{Question Relevance} \\
 \hline
  \textbf{Safurai-001} & 95 & 90 & 95 & 95 \\
  \textbf{ChatGPT} & 85 & 90 & 80 & 85 \\
\hline
\hline
\end{tabularx}
}
\end{table}
\par
Although the two responses are quite similar, as previously shown, GPT4-based MultiParameters Evaluation is able to identify small details and differences that HumanEval would not be able to identify given its quantitative nature.

\section{Conclusion}
This research reveals deep insights into how the utilization of data transformation can lead to the creation of complex, high-quality datasets. Following the stance of the Orca team (Gunesekar er al. [2023]), we stress the profound significance of the datasets chosen for fine-tuning. Through the use of innovative data transformation and prompt engineering techniques, these datasets can critically enhance the efficiency of foundation models. This, in turn, can lead to sizeable augmentations in classic benchmarks. 

Our study presented Safurai-001 as a potent alternative to WizardCoder [\cite{wizardcoder}] for multi-turn conversations, adeptly managing user interactions and providing supports where needed. We also present a distinct advance in the area of Coding LLMs evaluation tools, GPT4-based MultiParameter Evaluation Benchmark, moving beyond purely quantitative measures to offer more qualitative indicators. This new method provides a richer, more nuanced means of evaluation that can expose insights beyond the reach of traditional standards such as HumanEval\footnote{https://github.com/openai/human-eval} and MultiPL-E\footnote{https://huggingface.co/datasets/nuprl/MultiPL-E}. Drawing on these findings, we aim to shape future discourse and innovation in this field, accentuating the paramount importance of robust evaluation tools and the fullest utilization of datasets in optimizing language model performance.

\subsection{Safurai-001 and WizardCoder outcome comparison}

\begin{center}
    \textbf{Question}
\end{center}

\begin{tcolorbox}
How can I write a Python function that takes a list of integers as an input and returns a tuple representing the sum and product of all integers in that list? 
\end{tcolorbox}

\begin{center}
    \textbf{WizardCoder Response}
\end{center}

\begin{tcolorbox}
We can use the built-in `sum()` function in Python to get a sum of all integers in the list. 
We can also use a variable initialized as 1 for product and then multiply it with each integer in the list. 
Here's the solution: 
\begin{lstlisting}[language=Python, 
frame=single,
xleftmargin=1cm, 
xrightmargin=1cm]
from typing import List, Tuple
def sum_product(numbers):
    total_sum = sum(numbers) or 0 # if numbers is empty, set total_sum to 0
    total_product = 1
    for num in numbers:
        total_product *= num
    return (total_sum, total_product)
\end{lstlisting}
Here's how the function works step by step:
\begin{enumerate}
    \item We initialize `total\_sum` as sum of all integers in the list using built-in `sum()` or set it to 0 if the list is empty using short circuit evaluation (`or`).
    \item We also initialize `total\_product` variable as 1 and then multiply each integer in the list with this variable.
    \item Finally, we return a tuple of `(total\_sum, total\_product)`.
\end{enumerate}
\end{tcolorbox}
\newpage
\begin{center}
    \textbf{Safurai-001 Response}
\end{center}
\begin{tcolorbox}
First, we need to understand what a list is. A list is a collection of items, like a grocery list. In this case, our list is made up of numbers.

Next, we need to understand what a tuple is. A tuple is like a list, but once it's made, we can't change it. It's like a locked box that we can't open.

Now, let's look at the problem. We need to write a function that takes a list of numbers and returns a tuple with the sum and product of those numbers.
Here's the code:
\begin{lstlisting}[language=Python, 
frame=single,
xleftmargin=1cm, 
xrightmargin=1cm]
from typing import List, Tuple

def sum_product(numbers: List[int]) -> Tuple[int, int]:
    """
    For a given list of integers, return a tuple consisting of a sum and a product of all the integers in a list.
    Empty sum should be equal to 0 and empty product should be equal to 1.
    """
    # If the list is empty, return a tuple with a sum of 0 and a product of 1
    if not numbers:
        return 0, 1

    # Otherwise, calculate the sum and product
    sum = 0
    product = 1
    for num in numbers:
        sum += num
        product *= num

    # Return the tuple with the sum and product
    return sum, product
\end{lstlisting}

Let's break it down:

\begin{itemize}
    \item `def sum\_product(numbers: List[int]) -\texttt{>} Tuple[int, int]:` - This line defines our function and tells Python that it takes a list of integers as input and returns a tuple with two integers (the sum and product).
    \item `if not numbers:` - This line checks if the list is empty. If it is, we return a tuple with a sum of 0 and a product of 1.
    \item `sum = 0` and `product = 1` - These lines initialize variables for the sum and product.
    \item `for num in numbers:` - This line loops through each number in the list.
    \item `sum += num` and `product *= num` - These lines add the current number to the sum and multiply it by the current product.
    \item `return sum, product` - This line returns the tuple with the sum and product.
\end{itemize}

So, if we call our function with the list `[1, 2, 3, 4]`, it will return the tuple `(10, 24)`. The sum is 10 (1 + 2 + 3 + 4) and the product is 24 (1 x 2 x 3 x 4).

I hope that makes sense! Let me know if you have any questions.
\end{tcolorbox}

\newpage
\subsubsection*{Acknowledgments}

\bibliography{iclr2024_conference}
\bibliographystyle{iclr2024_conference}

\end{document}